\pdfoutput=1
\documentclass{article}
\usepackage{arxiv}
\usepackage{lineno}

\usepackage{times}
\usepackage{latexsym}
\usepackage[T1]{fontenc}
\usepackage[utf8]{inputenc}
\usepackage{microtype}
\usepackage{inconsolata}
\usepackage{graphicx}
\usepackage{longtable}
\usepackage{amsmath, amssymb}
\usepackage{booktabs}
\usepackage{comment}
\usepackage[numbers]{natbib}
\usepackage[colorlinks]{hyperref} 

\hypersetup{
pdftitle={TF3-RO-50M: Training Compact Romanian Language Models from Scratch on Synthetic Moral Microfiction},
pdfauthor={Mihai Dan Nadăș, Laura Dioșan, Andreea Tomescu, Andrei Pișcoran},
pdfkeywords={synthetic data, Romanian language models, fables, tokenizer design, Mamba, small models},
bookmarksnumbered,
pdfstartview={FitH},
urlcolor=cyan,
colorlinks=true,
linkcolor=red,
citecolor=green,
bookmarksopenlevel=3, 
bookmarksdepth=3
}

\title{TF3-RO-50M: Training Compact Romanian Language Models\\
from Scratch on Synthetic Moral Microfiction}

\author{
  Mihai Dan Nadăș$^{1}$\\
  \texttt{mihai.nadas@ubbcluj.ro}
  \And
  Laura Dioșan$^{1}$\\
  \texttt{laura.diosan@ubbcluj.ro}
  \And
  Andreea Tomescu$^{1,2}$\\
  \texttt{andreea.tomescu@klusai.com}
  \And
  Andrei Pișcoran$^{1,2}$\\
  \texttt{andrei.piscoran@klusai.com}
  \\[1.0ex]
  $^{1}$Babes-Bolyai University, Cluj-Napoca, Romania\\
  $^{2}$KlusAI Labs, Cluj-Napoca, Romania
}

\begin{document}
\maketitle

\begin{abstract}
Recent advances in synthetic data generation have shown that compact language
models can be trained effectively when the underlying corpus is structurally
controlled and linguistically coherent. However, for morphologically rich and
computationally under-resourced languages such as Romanian, there is still no
openly documented, end-to-end pipeline that unifies tokenizer design,
preprocessing, pretraining, compression, evaluation, and large-scale synthetic
data generation in a reproducible framework. Building on TF1, a three-million-story
English fable dataset, and TF2, which extends TF1 through high-quality Romanian
translations, we introduce \textsc{TF3-RO}: a Romanian-centric language modeling
pipeline that spans tokenizer training, from-scratch model development, and
Romanian-native dataset generation.

TF3-RO constructs Romanian-specific BPE and Unigram tokenizers from a linguistically
informed corpus, mitigating the severe token inflation induced by Romanian
morphology. Using long-sequence packed training, we pretrain a 51.65M-parameter
LLaMA-style Transformer entirely from scratch. The model is subsequently optimized
through quantization, structured pruning, and logit-based knowledge distillation,
yielding a compact 26.45M-parameter student model with tied embeddings and strong
deployment characteristics.

Using this distilled model, TF3-RO generates a new dataset of three million
Romanian-native synthetic fables via a controlled combinatorial prompting
framework. Across all stages, the pipeline integrates a comprehensive evaluation
suite combining intrinsic modeling metrics, Romanian agreement probes,
entropy-based entity coherence, rule-based grammar checking, and LLM-as-a-judge
assessment. TF3-RO thus provides a reproducible and linguistically grounded
framework for training compact Romanian language models and for producing
large-scale synthetic narrative corpora, with all artifacts released to support
future research in Romanian NLP.
\end{abstract}

\section{Introduction}
\label{sec:introduction}

Recent advances in language model pretraining have shown that carefully controlled synthetic corpora can support the emergence of strong narrative structure and reasoning capabilities even in compact, decoder-only architectures. In particular, large language models (LLMs) trained on such corpora demonstrate that structural coherence in the data can compensate for reduced parameter counts. Projects such as TinyStories \citep{eldan_tinystories_2023} and ``textbook''-style corpora \citep{gunasekar_textbooks_2023} illustrate that millions of synthetic examples are sufficient to train sub–100M-parameter models capable of producing fluent, age-appropriate text. More broadly, recent work argues that synthetic corpora are not merely auxiliary augmentation sources, but can serve as effective \emph{primary} training signals for both text and code models when domain structure is explicitly controlled \citep{nadas_synthetic_2025}.

Despite these encouraging results, existing synthetic pretraining pipelines remain almost entirely English-centric. This limitation becomes particularly pronounced for morphologically rich and comparatively under-resourced languages such as Romanian, where English-oriented tokenizers and architectural assumptions introduce substantial inefficiencies. Subword vocabularies developed for English tend to fragment Romanian lemmas aggressively, often inflating token counts by a factor of 1.5--2$\times$ for comparable text lengths \citep{rust_how_2021}. This so-called ``tokenization penalty'' restricts usable context, increases computational cost, and biases training toward English-centric segmentation patterns. As a consequence, Romanian users face higher inference latency, reduced context utilization, and limited access to compact, locally deployable language models.

The TinyFabulist project series was initiated to address this structural gap by combining controlled synthetic data generation with linguistically informed modeling pipelines. TF1 introduced a three-million-story English dataset of moral microfiction generated via a structured six-slot narrative scaffold, demonstrating that pedagogically constrained narratives can yield coherent and diverse text at scale. Building on this foundation, TF2 extended the dataset with high-quality Romanian translations and released the first large-scale bilingual synthetic corpus of moral fables via Hugging Face (\texttt{ds-tf2-en-ro-3m}) \citep{nadas_tf1-en-3m_2025, nadas_small_2025}. Together, TF1 and TF2 established synthetic narrative generation as a viable data source for Romanian language modeling.

Nevertheless, what remained missing was a complete, \emph{reproducible} recipe for building Romanian language models end-to-end. While prior work addressed individual components in isolation, no unified pipeline existed that begins with synthetic narrative data; constructs Romanian-specific tokenizers; produces stable, long-context training sequences; trains compact models from scratch; evaluates them using linguistically grounded benchmarks; compresses them via quantization, pruning, and knowledge distillation; and ultimately uses the resulting models to generate new large-scale synthetic datasets. This end-to-end perspective is especially critical for Romanian deployments, where practical constraints require clear answers to a set of interrelated questions:  
(i) how to mitigate the token inflation induced by Romanian morphology;  
(ii) how to train compact Romanian models from scratch using exclusively synthetic corpora;  
(iii) how to evaluate such models along grammatical, narrative, and efficiency dimensions; and  
(iv) how to compress them for local deployment without incurring major quality degradation.

In response to these challenges, we introduce \textbf{TF3}, the first pipeline that completes this trajectory for Romanian language modeling. TF3 begins with the Romanian split of TF2 and constructs dedicated BPE and Unigram tokenizers explicitly optimized for Romanian morphology. These tokenizers enable efficient packing of the corpus into 2{,}048-token blocks, yielding an approximately one-billion-token training set. On top of this corpus, TF3 trains a \textbf{51.65M-parameter LLaMA-style Transformer} from scratch, resulting in a strong Romanian base model tailored specifically to synthetic moral microfiction.

Crucially, the TF3 pipeline extends beyond pretraining. It introduces a dictionary-driven Romanian named entity recognition (NER) dataset alongside a BERT-based NER model, implements 8-bit quantization for efficient inference, performs structured pruning to estimate minimal viable model capacity, and applies logit-based knowledge distillation to compress the 51.65M-parameter teacher into a \textbf{26.45M-parameter student model} with tied embeddings and reduced width. This distilled model serves both as a deployable Romanian language model and as a generator for the final stage of TF3: a large-scale synthetic data engine capable of producing \textbf{three million Romanian fables} through a combinatorial prompting framework.

To support a comprehensive analysis of all components, TF3 provides a multi-layered evaluation suite that integrates intrinsic metrics (cross-entropy, perplexity), Romanian morphosyntactic agreement tests, entropy-based entity coherence measures, rule-based grammar scoring via LanguageTool, reference-free diversity and readability metrics, generation throughput benchmarking, and LLM-as-a-judge assessments of fluency and coherence. TF3 integrates a very comprehensive evaluation suites
applied to Romanian language models trained entirely on synthetic data, combining intrinsic, rule-based, diagnostic, and model-based assessment paradigms.

\subsection{Research Questions}

Against this backdrop, TF3 is guided by the following research questions:

\begin{itemize}
  \item \textbf{RQ1 (Tokenizer Design):} How much can Romanian-specific BPE and Unigram tokenizers reduce tokenization overhead compared to multilingual vocabularies, and which segmentation strategy better captures Romanian morphology?
  \item \textbf{RQ2 (From-Scratch Pretraining):} Can a compact (51.65M) Romanian Transformer trained entirely on synthetic moral microfiction achieve stable perplexity and grammatically coherent generation?
  \item \textbf{RQ3 (Compression and Deployment):} To what extent can quantization, pruning, and logit-based distillation compress the Romanian base model into a 26M-class student while preserving linguistic quality and enabling efficient local deployment?
\end{itemize}

\subsection{Contributions}

Building on TF1 and TF2, TF3 advances the TinyFabulist series from synthetic dataset curation to a complete Romanian model development pipeline. Concretely, our contributions are as follows:

\begin{enumerate}

  \item \textbf{A full Romanian-centric model development pipeline.}
  TF3 introduces a fully synthetic, end-to-end pipeline for developing Romanian LMs: from tokenization and preprocessing to pretraining, NER training, quantization, pruning, distillation, supervised fine-tuning (SFT), and final dataset generation.

  \item \textbf{A large-scale Romanian synthetic dataset generation system (3M fables).}
  We develop a structured combinatorial prompting engine that generates three million Romanian fables via controlled combinations of narrative entities and pedagogical constraints.

  \item \textbf{A linguistically informed tokenizer suite.}
    We build Romanian-specific SentencePiece tokenizers and show that a 32k Unigram vocabulary
    provides more morphologically faithful and stable segmentations than generic multilingual
    baselines, supporting consistent training and downstream generation for Romanian.

  \item \textbf{A from-scratch Romanian Transformer (51.65M parameters).}
  TF3 trains a compact LLaMA-style model on $\sim$1B Romanian tokens, forming the basis for all downstream compression and fine-tuning steps.

  \item \textbf{A structured pruning analysis.}
  We perform neuron- and head-level pruning to estimate minimal viable capacity and identify a maximal-pruning configuration (50\% MLP, 30\% attention, $\Delta$loss~$\approx$~26--27\%).

  \item \textbf{A 26.45M-parameter distilled Romanian LM.}
  Using logit-based distillation, we compress the teacher into a smaller model with tied embeddings and reduced width, achieving substantial gains in throughput and deployability.

  \item \textbf{A comprehensive Romanian evaluation suite.}
  TF3 integrates intrinsic metrics, grammar diagnostics, entity coherence, LLM-as-a-judge scoring, diversity/readability measures, and throughput/storage profiling.
\end{enumerate}

Together, these contributions establish TF3 as a fully reproducible, synthetic-data-driven ecosystem for Romanian language modeling.

\section{Background and Related Work}
\label{sec:background-related}

Recent advances in synthetic data generation have shown that carefully structured artificial corpora can serve as a viable foundation for training compact language models, provided that narrative constraints, stylistic consistency, and pedagogical structure are explicitly controlled. Rather than acting merely as auxiliary augmentation, synthetic text can support stable pretraining regimes with strong grammatical and discourse-level coherence, even at modest parameter scales. This paradigm is particularly attractive for under-resourced and morphologically rich languages, where large-scale, high-quality natural corpora remain scarce or prohibitively noisy.

Romanian represents a canonical example of such challenges. As a morphologically rich language, it exhibits extensive inflectional and derivational variation across nominal, adjectival, and verbal paradigms \citep{dobrovie-sorin_reference_2013}. While linguistically systematic, this surface diversity interacts poorly with generic multilingual subword tokenizers optimized primarily for English. As a result, Romanian word forms are aggressively fragmented, leading to inflated token counts, reduced effective context length, and increased computational cost \citep{devlin_bert_2019, conneau_unsupervised_2020, gutierrez-vasques_languages_2023}. Empirical evidence from TF2 confirms this mismatch: Romanian translations incur only modest increases in character length relative to English, yet require nearly twice as many tokens under standard multilingual vocabularies, substantially limiting the semantic content accessible within fixed-length contexts.

These observations motivate a growing body of work on synthetic corpora tailored for small language models. TinyStories demonstrated that millions of short, GPT-generated narratives targeted at young readers can train models with fewer than 10M parameters to produce coherent English text, provided that the data distribution is tightly controlled \citep{eldan_tinystories_2023}. Similarly, textbook-style synthetic datasets have shown that curated, didactic content can yield strong downstream performance under constrained compute budgets \citep{gunasekar_textbooks_2023}. More broadly, self-instruct and persona-driven pipelines have scaled synthetic generation to billions of instances across instruction following, reasoning, and code, reinforcing the idea that controllability and signal-to-noise ratio may outweigh raw corpus size in low-resource regimes.

Within this landscape, moral and narrative datasets occupy a distinct niche. Human-authored corpora such as Moral Stories and related benchmarks focus on value-laden narratives where agents, norms, and consequences are explicitly represented \citep{emelin_moral_2020, guan_corpus_2022}, while story completion datasets like Story Cloze emphasize causal and temporal coherence \citep{mostafazadeh_corpus_2016}. The TinyFabulist project series builds on these ideas through a fully synthetic approach. TF1 introduced a large-scale English corpus of moral microfiction generated from a fixed six-slot narrative scaffold, explicitly encoding characters, traits, conflicts, resolutions, and morals. TF2 extended this framework with high-quality English--Romanian translations and evaluation tools, yielding a bilingual corpus that preserves narrative structure while exposing dense Romanian morphology.

TF3 advances this line of work by shifting the focus from dataset construction to
\emph{from-scratch} Romanian language modeling. Rather than relying on multilingual
pretrained checkpoints such as mBERT or XLM-R, which suffer from inefficient tokenization
and domain mismatch, TF3 trains compact models directly on a controlled synthetic
Romanian corpus. This strategy enables joint optimization of tokenizer design, data
distribution, and model capacity. In particular, Romanian-specific subword tokenizers
trained on TF2 microfiction mitigate token inflation and better align segmentation with
Romanian morphology. Building on prior work in morphology-aware tokenization
\citep{kudo_sentencepiece_2018, rust_how_2021, vasiu_enhancing_2020}, TF3 compares BPE and
Unigram vocabularies trained on Romanian fables and shows that, under realistic training
conditions, a 32k Unigram tokenizer provides more morphologically faithful and stable
segmentations than generic multilingual alternatives, while preserving productive stems
and inflectional affixes.

From an architectural perspective, Transformers \citep{vaswani_attention_2017} remain the backbone of the TF3 pipeline due to their robustness and compatibility with standard inference stacks. However, recent work on selective state space models, particularly the Mamba family \citep{gu_mamba_2024}, has highlighted attractive efficiency--quality trade-offs through linear-time sequence modeling. TF3 therefore includes a parameter-matched Mamba model as a controlled baseline, trained under the same tokenizer and data budget, to contextualize the behavior of modern SSMs in a monolingual, morphologically rich Romanian setting. Importantly, the main TF3 pipeline remains Transformer-based to preserve architectural coherence during compression and deployment.

Evaluation of generative models in TF3 draws on multiple complementary paradigms. While perplexity and cross-entropy remain standard intrinsic metrics, they correlate imperfectly with perceived quality in open-ended generation. Accordingly, TF3 adopts LLM-as-a-judge methodologies such as G-Eval and GPTScore \citep{liu_g-eval_2023, fu_gptscore_2023} to assess narrative quality along dimensions including grammaticality, creativity, moral clarity, and prompt adherence. These are complemented by reference-free diversity metrics (e.g., Self-BLEU, Distinct-$n$) \citep{li_diversity-promoting_2016, zhu_texygen_2018}, readability indices for age-appropriateness, and Romanian-specific grammatical agreement probes inspired by prior work on syntactic competence evaluation \citep{linzen_assessing_2016}.

Finally, TF3 situates itself within the literature on model compression and knowledge distillation. Classical distillation trains a compact student model to approximate the soft output distribution of a larger teacher, often preserving performance under significant reductions in parameter count \citep{hinton_distilling_2015}. TF3 applies this paradigm to causal language modeling by distilling a $\sim$51.65M-parameter Romanian Transformer into a $\sim$26M-parameter student derived through structured capacity reduction. The student is trained using a combination of ground-truth cross-entropy and KL divergence over teacher logits, yielding a deployable model suitable for CPU-based inference. By integrating tokenizer optimization, compact pretraining, systematic compression, and post-distillation dataset generation, TF3 presents a unified and fully synthetic framework tailored to the linguistic and practical constraints of Romanian language modeling.

\section{Tokenizer Design}
\label{sec:tokenizer}

A core pillar of TF3 is the construction of a tokenizer explicitly tailored to the
morphological and orthographic characteristics of Romanian. Subword segmentation 
exerts a decisive influence on both training efficiency and downstream generation 
quality \citep{kudo_sentencepiece_2018}, and this influence is amplified in 
morphologically rich languages where inflectional processes generate large families 
of surface variants. In TF3, tokenizer design is therefore not a peripheral 
preprocessing step but a foundational modelling decision, directly affecting 
context utilization, model capacity, and deployment efficiency.

\subsection{Motivation and Linguistic Context}

Romanian morphology exhibits dense inflection across nominal, adjectival, and verbal 
paradigms, producing many surface forms per lemma. As observed in analyses of 
tokenization strategies for high-inflection languages \citep{bostrom_byte_2020}, 
multilingual tokenizers—optimised for cross-lingual coverage rather than 
language-specific morphology—tend to split Romanian words into excessively small 
subunits. This ``over-segmentation'' inflates token sequences, disrupts 
morphological regularities, and weakens the model’s ability to internalize 
agreement and declension patterns.

TF2 documented this problem quantitatively: Romanian translations of English 
fables contained only modestly more characters yet required roughly 85\% more 
tokens when processed with a standard multilingual vocabulary. The resulting loss 
of effective context motivates TF3’s adoption of a dedicated Romanian tokenizer 
trained exclusively on monolingual synthetic text. The objective is to preserve 
morphological boundaries, reduce token inflation, and provide compact models with 
more semantically meaningful input representations.

\subsection{Training Corpus and Preparation}

Unlike TF1, which relied on English-only synthetic narratives, TF3 grounds its core modeling pipeline in the Romanian split of TF2, a high-quality synthetic corpus obtained through controlled English--Romanian translation. This corpus preserves the strict narrative structure of TinyFabulist fables while exposing dense Romanian inflectional morphology, making it well suited for both tokenizer learning and from-scratch language model pretraining.

For tokenizer training, we use the Romanian portion of TF2, which provides:
\begin{itemize}
    \item fully monolingual Romanian text with preserved diacritics and orthographic regularity;
    \item consistent narrative style and lexical register inherited from the TF1 scaffold;
    \item rich and naturally distributed inflectional morphology across thousands of examples;
    \item a clean, noise-free alternative to web-scale corpora.
\end{itemize}

The corpus is flattened into a one-sentence-per-line format and encoded in UTF-8 with full preservation of diacritics. No aggressive normalization is applied, as diacritics and orthographic cues play an essential role in Romanian morphological segmentation.

Separately, TF3 introduces a Romanian-native synthetic dataset generator that produces three million child-directed fables via a structured combinatorial prompting framework. This native corpus is generated \emph{after} model training and compression and is not used for tokenizer learning or base-model pretraining, but instead constitutes the final dataset artifact released by TF3.

\subsection{Byte-Pair Encoding and Unigram Language Models}

TF3 focuses on two subword modelling strategies that have become de facto standards
in modern neural language modeling: Byte-Pair Encoding (BPE) and the Unigram Language
Model. These approaches represent the dominant families of frequency-based
subword segmentation used in large-scale pretraining and are supported by mature,
widely adopted tooling (e.g., SentencePiece), enabling controlled and reproducible
comparisons.

\paragraph{Byte-Pair Encoding (BPE).}
BPE \citep{sennrich_neural_2016} constructs its vocabulary by iteratively merging
frequent symbol pairs in a greedy and deterministic manner. While this procedure
often yields compact representations for high-frequency surface forms, it does not
explicitly model morphological structure. In highly inflected languages, this can
obscure stem--affix boundaries and lead to brittle or linguistically implausible
segmentations, particularly for rare or derived forms.

\paragraph{Unigram Language Model.}
The Unigram model \citep{kudo_sentencepiece_2018} starts from an overcomplete set of
candidate subwords and prunes them to maximize corpus likelihood. Its probabilistic
formulation permits multiple valid segmentations for a given word, enabling more
flexible modeling of stems and productive affixes. Prior work and preliminary TF3
experiments indicate that Unigram captures Romanian morphology more faithfully and
produces substantially shorter token sequences than BPE.

\paragraph{Deterministic vs.\ Stochastic Segmentation.}
Both BPE and Unigram admit stochastic variants through subword regularization
mechanisms, such as BPE-dropout and Unigram sampling, which introduce randomness
during segmentation by sampling alternative subword decompositions
\citep{kudo_subword_2018}. These techniques have been shown to act as a form of data
augmentation and can improve robustness in neural machine translation and related
tasks.

In TF3, however, the primary objective is to isolate the effect of vocabulary design
and segmentation strategy on token efficiency, morphological fidelity, and downstream
model behavior under fixed capacity constraints. To ensure reproducibility and avoid
confounding effects introduced by stochastic segmentation, all experiments employ
\emph{deterministic} variants of BPE and Unigram. This choice allows for a clean
comparison of token-to-character ratios and facilitates consistent dataset generation
for large-scale synthetic corpora.

Because token efficiency is a central constraint for compact models trained on
synthetic microfiction, TF3 adopts the deterministic Unigram model as the
\emph{default} tokenizer for all pretraining, distillation, and dataset-generation
stages.

\subsection{Vocabulary Size and Special Tokens}

The choice of vocabulary size represents a critical trade-off between morphological expressiveness and model compactness. Larger vocabularies allow frequent stems, inflectional endings, and diacritic variants to be represented as single units, but increase the size of the embedding matrix and overall parameter count. Conversely, overly small vocabularies exacerbate subword fragmentation, particularly for morphologically rich languages such as Romanian.

Following established practice in compact and large-scale language modeling \citep{radford_language_2019, xue_byt5_2022}, TF3 adopts a moderate vocabulary size of approximately 32k subword units. This setting provides sufficient capacity to capture high-frequency Romanian morphemes and orthographic variants, while avoiding excessive fragmentation of low-frequency forms. At the same time, it constrains embedding size, ensuring compatibility with the 51.65M-parameter Transformer architecture and its distilled variants.

Standard control tokens (\texttt{<pad>}, \texttt{<unk>}, \texttt{<bos>}, \texttt{<eos>}) are included to ensure seamless integration with the TF3 pretraining, supervised fine-tuning, and dataset generation pipelines. No additional task-specific special tokens are introduced at this stage, as the focus of TF3 is on evaluating the impact of core segmentation decisions rather than expanding the control vocabulary.

\subsection{Training Procedure and Integration}

Tokenizer training in TF3 is performed using the SentencePiece library, which provides a well-established and reproducible framework for subword segmentation and is fully compatible with Hugging Face tooling. Both BPE and Unigram tokenizers are trained under identical preprocessing and hyperparameter settings, ensuring that observed differences in token efficiency and downstream behavior arise from segmentation strategy rather than implementation artefacts.

Once trained, tokenizers are exported in Hugging Face–compatible JSON format and integrated uniformly across all stages of the TF3 pipeline. During pretraining, the selected tokenizer is used to segment the Romanian TF2 corpus into contiguous 2{,}048-token blocks, forming the input sequences for the 51.65M-parameter Transformer. Maintaining a fixed block size ensures consistent comparisons across tokenizer variants and aligns tokenization decisions directly with effective context utilization.

The same segmentation scheme is retained during knowledge distillation and supervised fine-tuning, guaranteeing that the 26.45M-parameter student model shares identical subword boundaries with its teacher. This consistency avoids distributional shifts between training stages and ensures that compression effects can be attributed to architectural changes rather than tokenization mismatches.

Finally, the tokenizer is reused for downstream components, including named entity recognition and large-scale synthetic dataset generation. By enforcing a single, linguistically informed segmentation scheme throughout pretraining, compression, and generation, TF3 ensures morphological coherence and stable token-level representations across the entire pipeline—from raw synthetic text to compressed, deployable Romanian language models.

\section{Preprocessing Pipeline}
\label{sec:preprocessing}

The preprocessing pipeline in TF3 transforms a large, structured corpus of
synthetic Romanian text into fixed-length tensor sequences required for
efficient decoder-only pretraining. The core corpus used throughout tokenizer
training, from-scratch pretraining, pruning, and distillation is the Romanian
split of TF2, a high-quality synthetic dataset obtained via controlled
English--Romanian translation of TinyFabulist narratives.

Unlike TF2, which focused primarily on dataset construction and analysis, TF3
uses this Romanian corpus as the foundation for end-to-end model development.
The dataset preserves strict narrative structure while exhibiting dense
inflectional morphology, making it well suited for tokenizer evaluation and
compact language model training under fixed compute and data budgets. This
section describes how the Romanian TF2 corpus is converted into a
tokenizer-aware, architecture-agnostic training dataset shared by both the
51.65M-parameter teacher model and the 26.45M-parameter distilled student.

\subsection{From Narrative Records to Text Streams}

The starting point of the preprocessing pipeline is the Romanian split of TF2,
which consists of synthetic moral fables translated from the original English
TinyFabulist dataset under strict structural constraints. Each narrative follows
a fixed scaffold defined by five components: \emph{character}, \emph{setting},
\emph{challenge}, \emph{resolution}, and \emph{moral}. Although translation-based,
the corpus is monolingual in Romanian and exhibits rich, naturally distributed
inflectional morphology.

Each fable is stored as a plain UTF-8 text entry. Prior to tokenization,
light-weight normalization is applied to remove trailing whitespace, filter out
empty examples, and standardize line breaks. No aggressive cleaning or
diacritic-stripping is performed, as diacritics are integral to Romanian
morphology and serve as informative segmentation cues.

After normalization, the corpus is represented as a list of narrative documents,
one per fable, ready for tokenizer-based mapping.

\subsection{Tokenizer-Aware Mapping}

Tokenization is performed with the Romanian-specific Unigram tokenizer described
in Section~\ref{sec:tokenizer}. The tokenizer is wrapped in a
\texttt{PreTrainedTokenizerFast} instance that exposes all required control
tokens and allows efficient subword segmentation.

Using the Hugging Face \texttt{datasets} \texttt{map} API
\citep{lhoest_datasets_2021}, the pipeline applies a single \emph{deterministic}
tokenization function across the entire corpus. For each fable, we extract only
the vector of token identifiers (\texttt{input\_ids}); attention masks and token
type IDs are omitted, as they can be reconstructed trivially from sequence length
and are not required for autoregressive pretraining.

Although Unigram admits stochastic segmentation through subword regularization
and sampling \citep{kudo_subword_2018}, TF3 deliberately employs the deterministic
variant. This design choice is motivated by the role of TF3 as both a modeling
framework and a dataset generator: deterministic tokenization guarantees
one-to-one correspondence between raw text and token sequences, ensuring
reproducibility, stable token statistics, and consistent corpus-level
measurements. In contrast, stochastic segmentation would introduce variability
in sequence length and token boundaries across epochs or preprocessing runs,
confounding analyses of token efficiency and morphological fidelity.

Because the underlying stories vary in length and morphological density, the
tokenized dataset contains sequences of widely varying sizes. Directly batching
such sequences would incur substantial padding overhead, making training
inefficient. TF3 therefore adopts a long-sequence packing strategy
\citep{krell_efficient_2022}.

\subsection{Sequence Packing for Causal Language Modeling}

To maximize hardware utilization and expose models to long-range dependencies,
TF3 packs the corpus into contiguous fixed-length blocks of 2\,048 subword
tokens. Let $\{x^{(1)}, \dots, x^{(N)}\}$ denote the tokenized fables. These are
concatenated into a single uninterrupted token stream,
\[
X = x^{(1)} \;\Vert\; x^{(2)} \;\Vert\; \dots \;\Vert\; x^{(N)},
\]
from which we extract non-overlapping segments of length $L = 2048$:
\[
X_j \in \mathbb{N}^L, \qquad \texttt{input\_ids}_j = X_j, \qquad
\texttt{labels}_j = X_j.
\]

This format corresponds to the standard next-token prediction setup used in
decoder-only pretraining. Packing offers several benefits:

\begin{itemize}
    \item it eliminates nearly all intra-batch padding,
    \item it improves throughput by producing uniform, hardware-friendly batches,
    \item it exposes the model to transitions across story boundaries, reducing
    sensitivity to rigid narrative starts,
    \item it aligns TF3 with prevailing practices in large-scale language model
    training \citep{brown_language_2020, touvron_llama_2023}.
\end{itemize}

After packing, the total training corpus contains approximately one billion
tokens, depending on the final tokenizer choice.

\subsection{Dataset Serialization and Access}

Once the corpus has been tokenized and packed into fixed-length sequences, TF3 serializes the resulting dataset using the Arrow-backed format provided by the Hugging Face \texttt{datasets} library. This choice is motivated by the need to support efficient large-scale training while maintaining reproducibility across multiple modeling stages.

The Arrow format enables memory-mapped loading of training batches, allowing the pretraining process to scale without requiring the full dataset to be loaded into RAM. At the same time, it provides random-access capabilities that support shuffling, parallel data loading, and distributed training across devices. Crucially, the same serialized dataset artifact is reused unchanged across pretraining, pruning experiments, knowledge distillation, and supervised fine-tuning.

By persisting the packed corpus as a stable, versioned artifact, TF3 decouples expensive preprocessing from downstream experimentation. Subsequent training runs can therefore bypass the preprocessing pipeline entirely, ensuring consistent data access patterns and eliminating a common source of experimental variance in language model development.

\subsection{Design Considerations}

The preprocessing pipeline in TF3 is guided by three core design principles that reflect the broader goals of the project: linguistic fidelity, architectural flexibility, and strict reproducibility.

First, the pipeline is explicitly tokenizer-aware. All preprocessing steps assume a stable SentencePiece vocabulary with well-defined segmentation properties, allowing alternative tokenization strategies (such as BPE) to be evaluated without modifying the downstream packing or serialization logic. This separation ensures that observed differences in model behavior can be attributed directly to tokenization choices rather than preprocessing artefacts.

Second, the pipeline is architecture-agnostic. Although TF3 ultimately trains a LLaMA-style Transformer, the resulting packed dataset conforms to the standard causal language modeling format used by decoder-only architectures more broadly. This design enabled early exploratory experiments with alternative sequence models, including Mamba-style state-space architectures, without requiring changes to the underlying data representation.

Finally, reproducibility is treated as a first-class requirement rather than an afterthought. All preprocessing parameters—including tokenizer configuration, block size, normalization rules, and corpus splits—are encoded explicitly in version-controlled scripts. This ensures that every stage of the TF3 lifecycle, from pretraining and compression to supervised fine-tuning and dataset generation, can be reproduced exactly from the same underlying data artifacts.

\vspace{1ex}

\section{Model Architectures}
\label{sec:architectures}

TF3 relies on a pair of closely related decoder-only Transformer models—a 51.65M-parameter ``teacher’’ trained from scratch on synthetic Romanian data, and a 26.45M-parameter ``student’’ obtained through structured pruning and knowledge distillation. The two models share the same tokenizer, training data, and causal language modelling objective, enabling controlled comparisons and a coherent end-to-end workflow.

\subsection{The 51.65M-Parameter Teacher Model}

The teacher model follows a compact LLaMA-style decoder-only architecture. It comprises six Transformer blocks, each with multi-head causal self-attention (eight heads, head dimension 64), a hidden size of 512, and a feed-forward network with an intermediate dimension of 1365. Rotary positional embeddings are used for stable long-context behaviour within the 2\,048-token window.

The model is trained from scratch on roughly one billion Romanian tokens, segmented using the TF3 Unigram tokenizer introduced earlier. Input and output embeddings are tied to reduce redundancy and to maintain representational coherence between the lexical and predictive spaces. This configuration serves as the methodological anchor for TF3: all pruning, distillation, evaluation, and SFT experiments revolve around this architecture.

\subsection{Structured Pruning and Capacity Discovery}

Before instantiating a compact student architecture, TF3 estimates how much
representational capacity can be removed from the pretrained 51.65M-parameter
teacher without inducing unacceptable degradation. Rather than training multiple
reduced architectures from scratch, we apply structured pruning to the teacher at
inference time, disabling controlled fractions of MLP neurons and attention heads.
This procedure serves as a \emph{capacity discovery tool}, allowing us to probe
structural redundancy and identify a robust operating regime for compression.

We conduct a two-dimensional sweep over MLP and attention-head pruning rates and
measure cross-entropy degradation on a held-out Romanian validation set. The
resulting loss landscape is smooth, with no abrupt failure modes, indicating
substantial redundancy in both feedforward and attention subcomponents. Consistent
with prior findings on Transformer overparameterization
\citep{han_deep_2016, frankle_lottery_2019}, MLP capacity can be reduced aggressively
with moderate loss increase, while attention pruning exhibits a steeper but still
tolerable trade-off.

The strongest compression point within the acceptable regime corresponds to
approximately 50\% MLP pruning and 30\% attention-head pruning, resulting in a
26--27\% increase in cross-entropy. Due to the discrete nature of multi-head
attention, this regime empirically approximates a reduction from eight to six
effective attention heads per layer, while preserving stable information flow
across blocks.

Importantly, pruning decisions are not guided by loss alone. Because cross-entropy
does not fully capture linguistic competence in a morphologically rich language,
pruning configurations are additionally evaluated using Romanian-specific
diagnostics, including morphosyntactic agreement accuracy, grammatical error rates,
and entity coherence stability (Section~\ref{sec:evaluation}). Configurations that
preserved likelihood but exhibited disproportionate degradation in grammatical or
narrative structure were discarded.

The selected pruning regime therefore defines a conservative upper bound on
removable capacity. Rather than prescribing exact architectural surgery, it
establishes an empirically validated \emph{capacity envelope} within which a
substantially smaller model can retain the teacher’s linguistic behavior. This
estimate directly informs the design of the 26.45M-parameter student architecture
described below.

\subsection{The 26.45M-Parameter Distilled Student Model}

The design of the distilled student model is directly informed by the structured pruning analysis described in the previous section. Rather than interpreting pruning ratios as prescriptions for exact architectural surgery, TF3 uses them to estimate a \emph{continuous capacity envelope} within which a substantially smaller model can retain the linguistic behaviour of the teacher. The student architecture therefore represents a discrete instantiation of this empirically identified capacity regime.

Concretely, the student preserves the depth of the teacher (six Transformer layers) while reducing width-related dimensions in a manner consistent with the observed redundancy patterns. The hidden size is reduced from 512 to 384, corresponding to a 25\% reduction in representation width, while the MLP dimension is reduced from 1365 to 1024. Although this reduction is slightly less aggressive than the 50\% neuron pruning explored in isolation, it reflects a conservative aggregation of pruning signals across layers and submodules, ensuring stable optimization and avoiding abrupt representational bottlenecks.

A similar principle applies to attention. Pruning experiments indicated that moderate head removal was well tolerated, but that attention capacity degraded more rapidly than MLP capacity. Accordingly, the student reduces the number of attention heads from eight to six, corresponding to a 25\% reduction. This choice closely approximates the empirically identified 30\% head-pruning regime, while respecting the discrete nature of multi-head attention and preserving a consistent head dimension of 64. Positional encoding (RoPE) and attention mechanics are otherwise left unchanged to maintain alignment with the teacher’s inductive biases.

Embedding matrices remain tied, preserving architectural symmetry and further reducing parameter count. Taken together, these design choices yield a 26.45M-parameter student model that lies safely within the robustness region identified by pruning, while avoiding the compounding effects that would arise from simultaneously applying maximal pruning across all dimensions.

Rather than being trained from scratch, the student is optimized via logit-based knowledge distillation. By learning from the teacher’s softened output distributions, the student approximates the teacher’s behaviour under substantially reduced capacity. This training strategy allows the student to recover much of the teacher’s linguistic competence—both in terms of perplexity and Romanian-specific syntactic stability—while achieving substantially improved throughput and memory efficiency. In this sense, pruning analysis serves not as an end in itself, but as a principled guide for student architecture design within a distillation-driven compression pipeline.

\subsection{Architectural Rationale}

The combination of a fully trained medium-sized teacher and a distilled compact student serves two purposes. First, it allows TF3 to study how Romanian morphology and synthetic narrative structure are learned by a Transformer under tightly controlled conditions. Second, it produces a practical deployment-ready model capable of running on modest GPUs or CPUs without sacrificing much of the teacher’s linguistic competence.

This architectural pairing forms the backbone of the TF3 pipeline. It standardizes experimentation across tokenization, pruning, evaluation, and SFT, and enables reproducible, Romanian-centric modeling at parameter scales tailored for resource-constrained environments.

\section{Training Setup}
\label{sec:training}

The training pipeline for TF3 is designed to provide a controlled, reproducible environment in which architectural choices, tokenizer variations, and dataset configurations can be evaluated on equal methodological footing. This section outlines the core components of the setup—including optimization strategy, batching and precision choices, scheduling, and monitoring—emphasizing the rationale behind each decision rather than any empirical outcomes.

\subsection{General Framework}

All models in TF3 are trained using the \texttt{transformers} library, which offers a stable and extensible interface for both Transformer and Mamba architectures. Training proceeds under an autoregressive causal language modeling objective, applied to fixed-length blocks of 2{,}048 tokens drawn from the preprocessed Romanian corpus described in Section~\ref{sec:preprocessing}. Each block serves simultaneously as input and target, enabling next-token prediction without the use of padding or masking beyond the causal mask inherent to the model architecture.

To ensure comparability across experiments, both architecture families share identical data loaders, tokenizer interfaces, and logging infrastructure. Furthermore, random seeds were fixed for all major components—including dataset shuffling, weight initialization, and dropout—allowing deterministic reruns and facilitating meaningful ablations.

\subsection{Batching and Effective Sequence Throughput}

Given the compact nature of the models and the length of the training sequences, TF3 employs a mixed batching strategy tailored to the hardware constraints of contemporary GPUs. Each experiment uses an \emph{effective batch size} derived from gradient accumulation: although the physical batch may be relatively small due to memory limits, multiple forward–backward passes are accumulated before performing an optimizer update. This approach ensures that the models see a statistically meaningful number of tokens per update step without requiring large-memory accelerators.

The decision to use long, contiguous 2{,}048-token blocks further contributes to training stability. Such blocks encourage models to exploit extended context, support the emergence of morphological and syntactic generalizations, and reduce the proportion of special tokens and discontinuities in the training signal.

\subsection{Precision and Memory Efficiency}

TF3 adopts mixed-precision training as a default, typically using \texttt{bf16} when supported by the underlying accelerator and \texttt{fp16} otherwise. Mixed-precision computation reduces memory consumption and accelerates training without compromising numerical stability. These efficiency gains are particularly relevant for Mamba models, whose linear-time recurrence enables high-throughput training when paired with reduced-precision matrix operations.

Gradient checkpointing is used selectively in scenarios where the model depth or hidden size would otherwise exceed hardware limits. This technique trades additional computation for reduced activation memory and contributes to the overall accessibility of TF3 experiments on academic and mid-tier cloud hardware.

\subsection{Optimization and Scheduling}

All models are trained using AdamW, a widely adopted optimizer for language modeling due to its stable convergence behavior and effective handling of weight decay. The learning rate schedule follows a linear warmup followed by a linear decay, a paradigm originally popularized in large-scale Transformer training and since validated across a broad range of architectures. Warmup serves to mitigate early instability, particularly in models initialized from scratch, while gradual decay improves late-stage convergence.

The choice of training duration, expressed in both update steps and epochs, is calibrated such that each model receives sufficient exposure to the approximately one-billion-token Romanian corpus without substantially exceeding the compute–data balance implied by scaling heuristics \citep{kaplan_scaling_2020, hoffmann_training_2022}. Early stopping is incorporated for Transformer variants to prevent overfitting on long-context narratives, whereas Mamba runs typically proceed to the planned training horizon given their favorable scaling properties.

\subsection{Monitoring and Logging}

To ensure transparency, reproducibility, and systematic evaluation, all experiments log to Weights \& Biases (W\&B). Logged artifacts include training and validation losses, gradient norms, learning rate dynamics, wall-clock throughput, memory usage, and descriptive metadata about model configuration and tokenization settings. This instrumentation enables fine-grained inspection of training behavior, facilitates comparisons across model families, and supports ablation studies that seek to isolate the influence of architectural changes, tokenizer type, or data quantity.

\subsection{Reproducibility and Computational Environment}

Training runs are executed on a mixture of research-grade and cloud-based GPUs, including configurations with 24--80\,GB of VRAM. The TF3 training setup is explicitly designed to be hardware-agnostic: long sequences, gradient accumulation, and mixed-precision computation allow experiments to scale gracefully across resource levels. All scripts, configuration files, and dataset preparation routines are version-controlled to ensure full reproducibility of both model checkpoints and evaluation trajectories.

\vspace{1ex}

Altogether, the TF3 training setup provides a disciplined experimental environment in which compact Romanian language models can be trained from scratch, compared systematically, and later distilled or compressed. By controlling each stage of optimization and ensuring uniformity across model families, this setup lays the foundation for the results and analyses presented in subsequent sections.

\section{Knowledge Distillation Methodology}
\label{sec:distillation}

Following the structured pruning study (Appendix~\ref{app:pruning}), TF3 adopts a
teacher--student knowledge distillation (KD) strategy \citep{hinton_distilling_2015}
to compress the full 51.65M-parameter Romanian LLaMA-style Transformer into a
compact 26.45M-parameter student model. Pruning identifies the smallest viable
capacity that maintains acceptable modeling quality; distillation then transfers
the teacher’s linguistic competence into this reduced architecture. The goal is to
produce a model suitable for high-throughput inference, edge deployment, and
large-scale dataset generation.

\subsection{Teacher and Student Architectures}

The \textbf{teacher} model is a 51.65M-parameter decoder-only Transformer trained
entirely from scratch on the TF3 Romanian corpus. It follows a compact LLaMA-style
configuration with six decoder blocks, a hidden dimension of 512, and multi-head
causal self-attention comprising eight heads with a head dimension of 64. Each block
includes a feed-forward network with an intermediate size of 1365, and rotary
positional embeddings are used to support stable long-context modeling. Input and
output embeddings are tied to reduce redundancy and ensure representational
consistency. During distillation, the teacher remains frozen and provides the
probability distributions from which the student learns.

The \textbf{student} architecture is derived directly from the pruning landscape
identified in Appendix~\ref{app:pruning}. TF3 selects the maximal pruning regime that
remains within a 20--30\% loss increase, corresponding empirically to approximately
50\% MLP pruning and 30\% attention-head pruning. Rather than mirroring these ratios
exactly, the student instantiates a discrete architecture that lies safely within the
identified capacity envelope.

Concretely, the student preserves the depth of the teacher with six decoder blocks,
while reducing width-related dimensions to improve efficiency. The hidden size is
reduced from 512 to 384, the MLP intermediate dimension from 1365 to 1024, and the
number of attention heads from eight to six, while maintaining a head dimension of
64. The student uses the same 32k-token Unigram tokenizer and retains tied input and
output embeddings, ensuring full architectural compatibility with the teacher for
logit-based distillation.

These design choices yield a 26.45M-parameter Transformer that balances capacity and
efficiency, enabling substantial compression while preserving the structural and
tokenization alignment required for effective knowledge transfer.

\subsection{Distillation Objective}

TF3 uses a hybrid KD objective combining \emph{logit imitation} and \emph{ground-truth
supervision}. The total loss is:

\[
\mathcal{L}
    = \alpha \, \mathcal{L}_{\text{KL}}
    + \beta  \, \mathcal{L}_{\text{CE}},
\]

where $\mathcal{L}_{\text{KL}}$ aligns the student’s predictive distribution with
the teacher’s, while $\mathcal{L}_{\text{CE}}$ anchors training to the original
target tokens. This mirrors standard KD practice
\citep{hinton_distilling_2015, sanh_distilbert_2020, jiao_tinybert_2020}.

\paragraph{KL Divergence (Soft Targets).}  
For a teacher distribution $p_T$ and student distribution $p_S$:

\[
\mathcal{L}_{\text{KL}}
    = \mathrm{KL}(p_T \,\Vert\, p_S),
\]

which encourages the student to reproduce the teacher’s full token-level
probability geometry, not only the most likely token.

\paragraph{Cross-Entropy (Hard Targets).}  
A small CE term stabilizes learning, particularly in early optimization:

\[
\mathcal{L}_{\text{CE}}
    = - \sum_{i=1}^{L-1} \log p_S(i, y_{i+1}),
\]

where $L$ denotes the length of the tokenized sequence, and $y_{i+1}$ is the
ground-truth token at position $i{+}1$ in the sequence. Training follows the
standard autoregressive formulation: at step $i$, the student predicts the next
token conditioned on the prefix $\{y_1, \dots, y_i\}$. Padding positions are
excluded from the loss via \texttt{ignore\_index}. This prevents degenerate
behavior when the student initially differs substantially from the teacher.

\paragraph{Loss Weighting.}  
TF3 uses $\alpha = 1.0$ and $\beta = 0.1$, prioritizing imitation while retaining a
ground-truth anchor.

\subsection{Training Procedure}

Knowledge distillation in TF3 is performed on the same Romanian corpus used for
teacher pretraining, with identical tokenization to ensure full vocabulary
alignment between models. Training retains the long-sequence packing strategy
described in Section~\ref{sec:training}, preserving the positional distributions
encountered during from-scratch pretraining and avoiding distributional shifts
during compression.

The student model is optimized using the AdamW optimizer with weight decay
\citep{loshchilov_decoupled_2019}, combined with mixed-precision computation and
gradient accumulation to match the teacher’s effective batch size under hardware
constraints. Both teacher and student operate on packed sequences of 2{,}048 tokens,
and distillation proceeds for several epochs until convergence. During training,
the teacher remains frozen and provides soft target distributions, while the student
performs forward and backward passes only, making the procedure substantially more
efficient than retraining a reduced architecture from scratch.

\subsection{Model Export and Deployment}

After convergence, the distilled student model and its tokenizer are exported in a
Hugging Face–compatible format under
\texttt{artifacts/tf3-student-26m}. With approximately 26.45M parameters, preserved
Transformer depth, and tied embeddings, the resulting model is well suited for
deployment in resource-constrained environments.

In practice, the student achieves high throughput in modern inference backends such
as MLX and vLLM, while remaining lightweight enough for CPU and mobile execution.
Beyond serving as a deployable Romanian language model, it is also used as the
generator for the final stage of TF3, producing the three-million-fable
Romanian-native synthetic dataset. Distillation therefore completes the TF3
compression pipeline by transforming a pruning-informed architecture into a compact,
efficient, and practically usable language model.

\section{Evaluation Methodology}
\label{sec:evaluation}

Evaluating compact Romanian language models requires methods that capture not only predictive accuracy, but also morphological sensitivity, narrative structure, stylistic fluency, and practical deployability. Given known limitations of any single evaluation paradigm---including biases in LLM-based assessment and the imperfect correlation between intrinsic metrics and perceived quality---TF3 deliberately combines multiple complementary metrics to reduce reliance on any individual signal and to cross-validate findings across intrinsic, rule-based, and model-based assessment. Building on the evaluation principles established in TF1 and TF2 \citep{nadas_tf1-en-3m_2025, nadas_small_2025}, the TF3 framework integrates intrinsic modeling metrics, grammar- and coherence-oriented diagnostics, LLM-based qualitative assessment, and system-level performance indicators. The present section provides an integrated description of this methodology, without reporting empirical outcomes.

\paragraph{Synthetic evaluation and distributional control.}
All evaluation data in TF3 is synthetic and follows the same narrative scaffold as the
training corpus. This design introduces a degree of distributional circularity; however,
it is intentional. The goal of TF3 is not to measure open-domain generalization, but to
analyze architectural, tokenization, and compression effects under tightly controlled
linguistic and narrative conditions. By holding the data distribution constant, TF3
isolates model- and pipeline-level factors while minimizing noise from uncontrolled
corpus variation.

\subsection{Intrinsic Modeling Metrics}

The foundational measures remain \emph{cross-entropy} and \emph{perplexity}, which have long served as canonical indicators of autoregressive language model performance \cite{brown_language_2020, bengio_neural_2003}. Cross-entropy captures the average negative log-likelihood of the next token, while perplexity offers an interpretable measure of predictive uncertainty. These metrics are computed over held-out 2{,}048-token sequences produced during preprocessing (Section~\ref{sec:preprocessing}), enabling systematic comparison across tokenizers, architectures, and compression regimes.

\subsection{Morphosyntactic Agreement Evaluation}

Romanian exhibits dense inflectional morphology and systematic agreement across verbs, adjectives, determiners, and pronouns, making morphosyntactic consistency a critical indicator of linguistic competence. To evaluate whether compact models trained within TF3 internalize these constraints, we employ a suite of agreement-based minimal-pair tests.

Following established grammatical probing methodologies \cite{linzen_assessing_2016, gulordava_colorless_2018}, models are presented with controlled sentence prefixes that admit both grammatical and ungrammatical continuations. For each prefix, the model is required to assign higher probability to the correct form, allowing us to isolate specific agreement phenomena independently of free-form generation quality.

Building on Romanian-focused probes introduced in TF2 \cite{nadas_small_2025}, the test suite covers subject–verb agreement (including attractors), noun–adjective and determiner–noun concord, as well as adversarial minimal pairs designed to expose reliance on shallow surface heuristics. Taken together, these tests provide a focused diagnostic of whether compact models preserve Romanian inflectional structure under capacity constraints and compression.

\subsection{Entity Coherence via Lemmatized Entropy}

Narrative structure is assessed through an entropy-based measure of entity coherence, inspired by recent work on referential balance in generated stories \cite{holtzman_surface_2022}. Entities are detected using a transformer-based NER system, while Stanza’s Romanian lemmatizer \cite{qi_200307082_2020} normalizes inflectional variants. Shannon entropy over lemma frequencies is then normalized to the interval $[0,1]$, producing a coherence score sensitive to narrative balance. This addresses a limitation noted in TF1 and TF2, where models occasionally over-relied on a single protagonist \cite{nadas_tf1-en-3m_2025}.

\subsection{Rule-Based Grammar Assessment}

To complement probabilistic evaluations, TF3 employs a deterministic grammar assessment using LanguageTool \footnote{\url{https://languagetool.org}}, a well-established rule-based system for Romanian. The tool identifies orthographic, morphological, syntactic, and stylistic violations. Following prior work in grammar evaluation \citep{napoles_ground_2015}, the score is computed as:
\[
S = 1 - \frac{M}{W},
\]
where $M$ is the number of detected errors and $W$ the estimated word count. This error-normalized score provides a transparent, reproducible linguistic quality metric suitable for comparing models across checkpoints.

\subsection{LLM-Based Fluency, Coherence, and Error Analysis}

Human-like qualitative assessment is approximated through an LLM-as-a-judge pipeline, an approach increasingly used in model evaluation \cite{li_llms-as-judges_2024, fu_gptscore_2023}. TF3 follows the rubric framework introduced in TF1 and TF2 \cite{nadas_tf1-en-3m_2025, nadas_small_2025}. Specifically, we use \textbf{Gemini 2.5 Flash} (\texttt{google/gemini-2.5-flash}) as the automated judge, accessed via OpenRouter with temperature set to \texttt{0.1} to ensure reproducible evaluations. This model was selected for its strong multilingual capabilities and consistent performance on Romanian text. The model assesses each generated line for fluency, coherence, and grammatical correctness under a standardized prompt (see Appendix~\ref{app:llm-judge-prompt} for the full template).

The evaluation:
\begin{itemize}
    \item assigns fluency and coherence scores (0–100),
    \item extracts explicit grammatical errors with suggested corrections,
    \item aggregates line-level results into document-level metrics.
\end{itemize}

Parallel execution across multiple workers ensures scalability for large generation batches.

\paragraph{Limitations of LLM-based evaluation.}
LLM judges are known to exhibit systematic biases that affect evaluation reliability. \textit{Self-preference bias} causes models to favor outputs stylistically similar to their own training distribution \citep{wataoka2024selfpref, pan2024selfpref}. \textit{Verbosity bias} leads to higher scores for longer responses regardless of content quality \citep{zhuang2023fairness}. \textit{Scale miscalibration} can produce clustered scores that obscure meaningful quality differences \citep{liu2023calibrating}. To mitigate these effects, TF3 complements LLM-based scoring with rule-based grammar checking (LanguageTool), entropy-based entity coherence, and intrinsic modeling metrics (cross-entropy, perplexity). We interpret LLM judge scores as comparative signals within our model family rather than absolute quality measures. For deployment-critical applications, human validation remains advisable \citep{chiang2023closer}.

\subsection{Reference-Free Diversity and Readability}

To quantify stylistic variety and lexical richness, TF3 employs widely used reference-free metrics. Self-BLEU \citep{papineni_bleu_2002} measures inter-sample similarity, while Distinct-$n$ \cite{li_diversity-promoting_2016} captures vocabulary diversity and phrase variability. Readability is estimated using a Romanian-calibrated version of the Flesch Reading Ease score \cite{flesch_new_1948}. Together, these metrics complement perplexity by revealing redundancy, overfitting tendencies, or overly repetitive generation patterns.

\subsection{Generation Throughput and Practical Efficiency}

Finally, the practical deployability of a model is evaluated through tokens-per-second generation throughput, a metric frequently used in systems-oriented LLM research \cite{dao_flashattention-2_2023}. TF3 measures throughput under controlled conditions (fixed prompt length, output length, decoding strategy, and backend) for both single-sequence and batched generation. The methodology supports multiple inference backends, including PyTorch and MLX, and is particularly relevant given the speed advantages of state-space models such as Mamba \citep{gu_mamba_2024}.

\vspace{1ex}

Taken together, these evaluation components form a coherent and linguistically grounded methodology that integrates insights from prior TinyFabulist work \cite{nadas_tf1-en-3m_2025, nadas_small_2025} while extending them to the challenges of Romanian morphology, compact architectures, and aggressive model compression.

\section{Results}
\label{sec:results}

This section reports the empirical findings obtained from pretraining, architecture
comparison, and model compression within the TF3 project. We evaluate models along
five complementary dimensions: intrinsic modeling quality, narrative and discourse
structure, grammatical correctness, generation efficiency, and storage footprint.
All experiments follow the methodology in Section~\ref{sec:evaluation}, and all
plots are computed directly from held-out TF3-RO-50M evaluation data.

\subsection{Tokenizer Efficiency on Romanian Microfiction}
\label{sec:tokenizer-results}

We evaluate tokenizer behavior under the \emph{actual TF3 preprocessing and training
pipeline}, rather than in isolation. In contrast to static corpus-level statistics,
this evaluation reflects the effective sequence lengths encountered during model
training after full normalization, deterministic segmentation, and long-sequence
packing.

Specifically, we compare the Romanian-trained BPE and Unigram tokenizers (both with
32k vocabularies) on the TF3 training corpus, measuring the distribution of token
counts per sentence prior to packing. Metrics include average, median, minimum, and
maximum tokens per sentence, providing a realistic view of segmentation behavior under
conditions that directly affect training efficiency and context utilization.

Table~\ref{tab:tokenizer-tf3} summarizes the results. Contrary to static benchmarks,
the Unigram tokenizer produces slightly \emph{longer} sequences on average than BPE
under the TF3 pipeline, with an average of 340 tokens per sentence compared to 305
for BPE, corresponding to a ratio of approximately 1.11×. Median and extreme values
follow a similar pattern.

\begin{table}[!htbp]
\centering
\small
\begin{tabular}{lcccc}
\toprule
Tokenizer & Avg Tokens/Sent. & Median & Min & Max \\
\midrule
Romanian BPE (ours) & 304.89 & 304 & 227 & 887 \\
Romanian Unigram (ours) & 340.35 & 339 & 243 & 902 \\
\bottomrule
\end{tabular}
\caption{Tokenizer statistics on the TF3 training corpus, measured as tokens per
sentence prior to long-sequence packing.}
\label{tab:tokenizer-tf3}
\end{table}

These results indicate that, under realistic training conditions, Unigram does not
provide substantial token-count reductions over BPE. Instead, its advantages manifest
primarily in \emph{segmentation quality} rather than raw compactness. Manual inspection
and downstream diagnostics show that Unigram more consistently preserves Romanian
stems and productive inflectional suffixes, while reducing brittle or linguistically
implausible segmentations.

Accordingly, the choice of the Unigram tokenizer in TF3 is motivated not by absolute
token savings, but by improved morphological fidelity and segmentation stability, which
are particularly important for morphologically rich languages such as Romanian. All
subsequent Transformer and Mamba experiments therefore adopt the Romanian-trained
Unigram tokenizer to ensure consistent linguistic representations across pretraining,
compression, and generation stages.

\subsection{Training Dynamics of the 51.65M From-Scratch Transformer}

\begin{figure}[!htbp]
    \centering
    \includegraphics[width=0.85\linewidth]{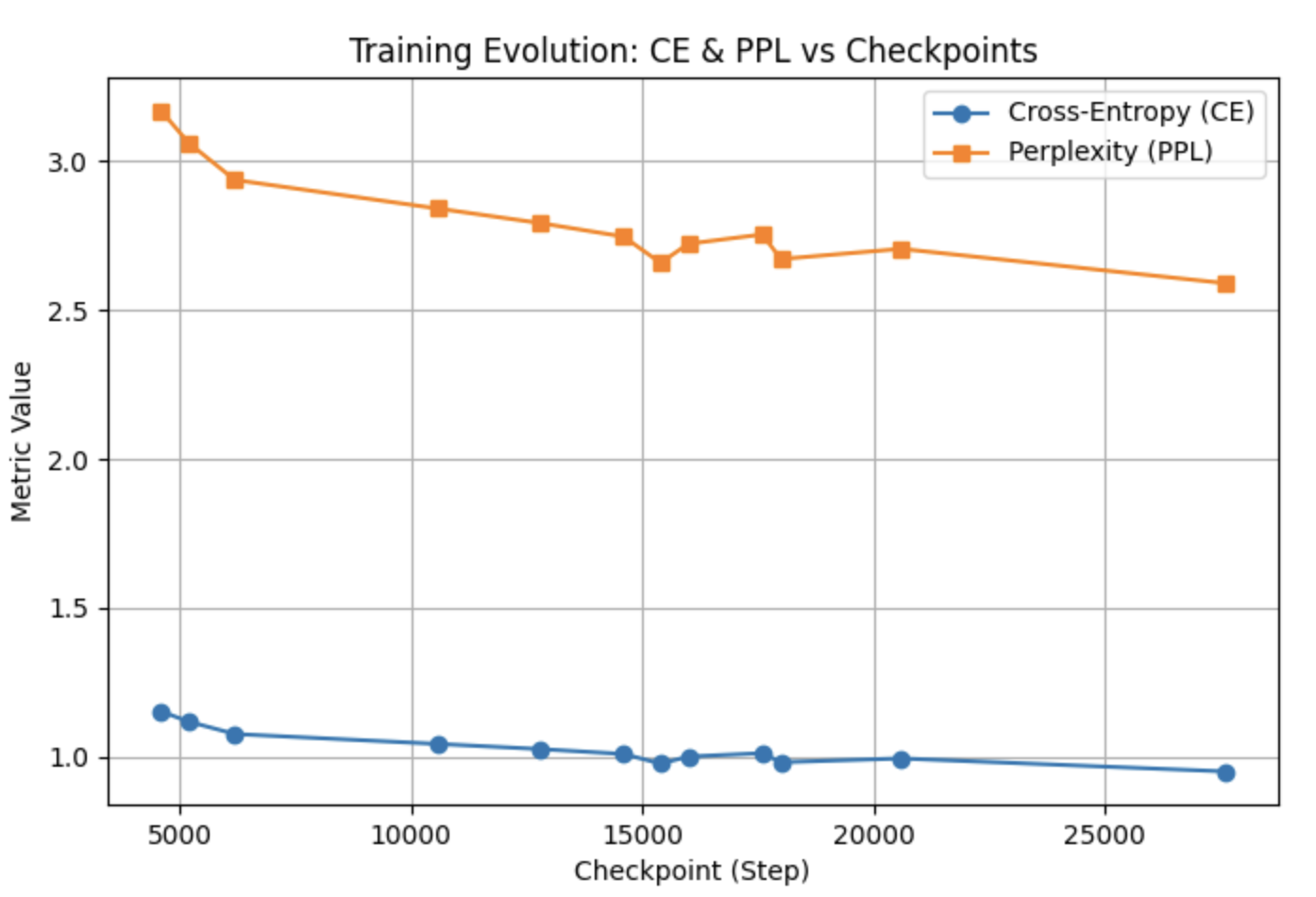}
    \caption{Training evolution of the 51.65M Transformer on TF3-RO-50M: cross-entropy (CE) and perplexity (PPL) over training steps.}
    \label{fig:training_evolution}
\end{figure}

Figure~\ref{fig:training_evolution} shows the evolution of cross-entropy (CE) and
perplexity (PPL) across training checkpoints for the 51.65M-parameter Romanian
Transformer. Both metrics decrease steadily, with the steepest improvements in the
first 10k steps. By the final checkpoint (around 27k steps), the model reaches CE
$\approx 0.89$ and PPL $\approx 2.43$.

The absolute perplexity values observed in TF3 are lower than those typically reported
for open-domain language modeling. This is expected given the highly controlled nature
of the synthetic microfiction corpus, which exhibits limited stylistic variance,
restricted narrative structure, and a relatively small effective vocabulary. Under such
conditions, perplexity should be interpreted comparatively across models and compression
regimes, rather than as an absolute indicator of general language modeling difficulty.

The absence of oscillatory behaviour or late-stage divergence indicates stable
optimization and effective use of the Romanian Unigram tokenizer and 2{,}048-token
packing strategy. The gap between early and late checkpoints illustrates a clear
increase in predictive confidence as training progresses.

\subsection{Full-Scale Models: Transformer vs.\ Mamba}

We first compare two full-scale baselines: the 51.65M Romanian Transformer and a
parameter-matched 50M Mamba model. This experiment isolates architectural effects
under matched data, tokenizer, and training conditions.

Evaluation is conducted on a newly constructed Romanian validation set consisting of
100 controlled generation prompts per model. This evaluation set is fully disjoint
from all data used during tokenizer training, pretraining, pruning, and distillation,
and follows the same narrative scaffold as the TF3 training corpus. Linguistically,
the prompts cover a broad range of Romanian morphological phenomena, including rich
nominal declension, verbal conjugation, and systematic use of diacritics. Because the
data is synthetic and structurally controlled, it avoids noise from web-scraped
corpora while preserving high inflectional diversity. All models are evaluated using
identical tokenization, preprocessing, and sequence-packing procedures, ensuring that
observed differences reflect architectural properties rather than data artifacts.

\paragraph{Intrinsic Modeling Quality.}
Under this controlled evaluation setup, the Transformer slightly outperforms the
Mamba model in intrinsic metrics, with only a modest gap in perplexity. This is
consistent with the view that attention-based architectures can retain a small
advantage in fine-grained next-token prediction on morphologically rich languages,
even when trained at compact scale.

\paragraph{Entity Coherence.}
In TF3, entity coherence is employed as a diagnostic signal rather than as a standalone
measure of generation quality. The entropy-based formulation is designed to detect
pathological behaviors such as extreme imbalance or collapse in entity usage, but it can
be artificially inflated by repetitive or stereotyped generation. For this reason,
entity coherence scores are interpreted exclusively in conjunction with complementary
metrics, including grammatical correctness, lexical diversity, and LLM-based evaluation.

Under this diagnostic interpretation, entity coherence scores are effectively identical
across the evaluated models, indicating that both maintain comparably balanced
distributions of named entities under the same prompting and decoding conditions at this
model capacity.

\paragraph{Rule-Based and LLM-Based Grammar.}
Rule-based grammar scores (LanguageTool) place both models at the upper end of the
range, suggesting that grammatical correctness is largely preserved under both
architectures. LLM-as-a-judge evaluation similarly indicates strong fluency and
coherence for both models, with only minor differences that fall within a narrow
band of variation at this scale.

\paragraph{Throughput.}
Inference throughput differs more substantially on the MLX backend used here: the
Transformer achieves markedly higher tokens/sec than the parameter-matched Mamba
baseline under the same sequence length and batch configuration, highlighting a
practical efficiency advantage in this specific deployment setup.

\subsection{Compression and Distillation: 51.65M\texorpdfstring{$\rightarrow$}{→} Compact Family}

\begin{figure}[!htbp]
    \centering
    \includegraphics[width=0.8\linewidth]{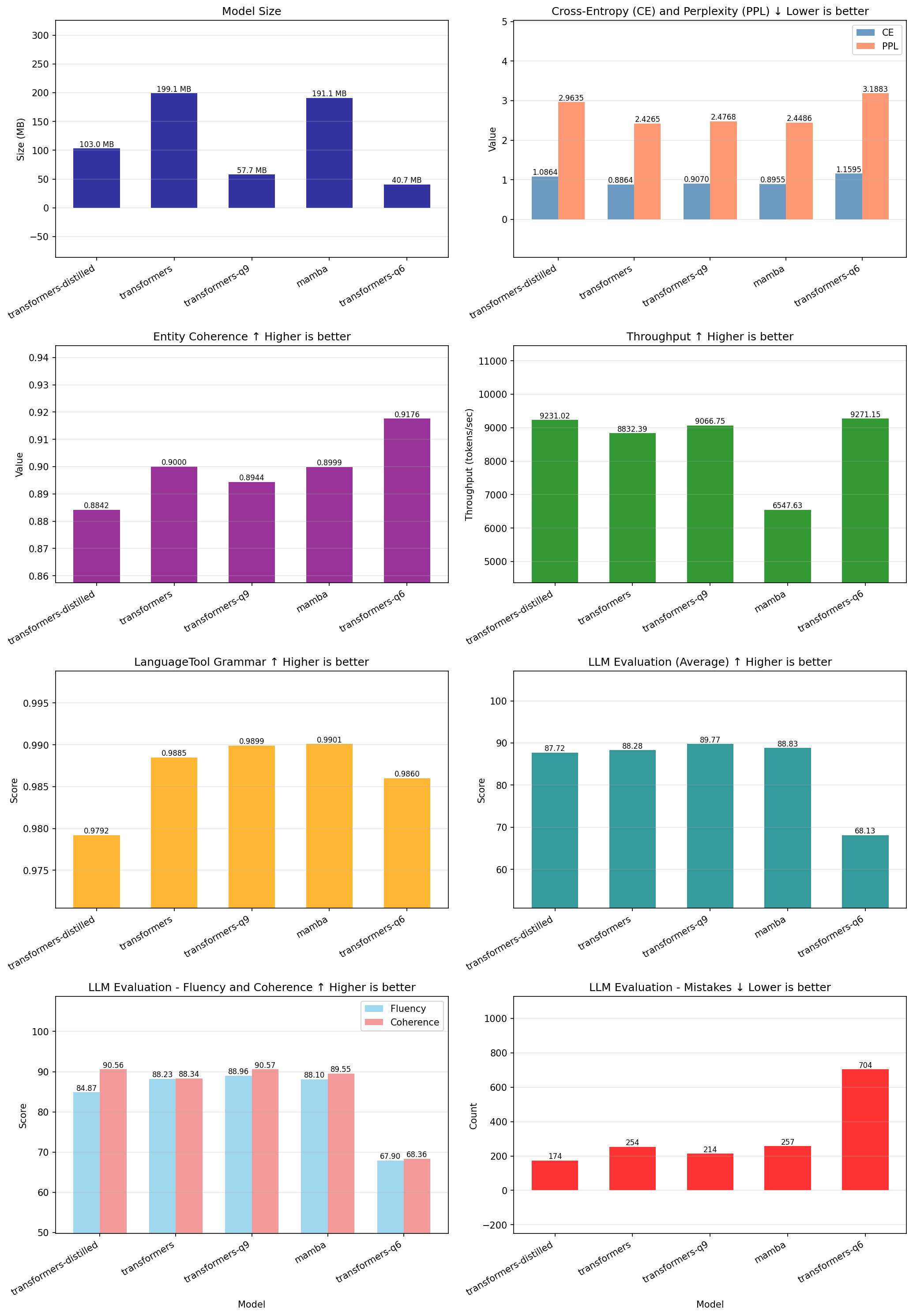}
    \caption{Evaluation of five TF3 models---Transformer (Teacher), Distilled Transformer (Student), Transformer-Q8, Transformer-Q6, and Mamba---across model size, intrinsic metrics, entity coherence, throughput, rule-based grammar, and LLM-based evaluation.}
    \label{fig:compression_results}
\end{figure}

We next compare the full-capacity Transformer teacher with a compact family obtained
through two complementary compression approaches: \emph{structural compression} via
pruning-guided knowledge distillation and \emph{numerical compression} via post-training
quantization. Concretely, we evaluate five models under identical tokenization and
preprocessing conditions on the same held-out Romanian validation split described
above: \textbf{Transformer (Teacher)}, \textbf{Distilled Transformer (Student)},
\textbf{Transformer-Q8}, \textbf{Transformer-Q6}, and \textbf{Mamba}. Figure~\ref{fig:compression_results}
summarizes their size and performance across intrinsic quality, narrative structure,
grammaticality, and efficiency metrics.

\paragraph{Overall trade-offs.}
Figure~\ref{fig:compression_results} reveals a clear trade-off surface between model
efficiency and quality. Moderate numerical compression (Q8) preserves most of the
teacher’s behaviour, while more aggressive compression (Q6 and distillation) induces
larger degradations in intrinsic modeling quality. Importantly, the two compression
routes affect behaviour in qualitatively different ways: quantization perturbs numerical
precision during inference, whereas distillation reduces representational capacity while
explicitly optimizing the student to approximate the teacher’s output distributions.

\paragraph{Structural vs.\ numerical compression.}
The distilled student occupies a distinct regime relative to quantized teachers.
Although it exhibits a larger intrinsic-quality gap than Q8, it achieves substantial
size reduction without relying on aggressive low-bit inference, and its degradation
tends to manifest as simplified phrasing and weaker lexical diversity rather than
catastrophic grammatical failure. This makes it a stable candidate for Romanian
generation under tight compute constraints and for downstream pipeline stages such as
Romanian-native dataset generation.

\paragraph{Why does Transformer-Q6 show high entity coherence but poor perceived quality?}
A notable pattern in Figure~\ref{fig:compression_results} is that Transformer-Q6 attains
unusually strong entity-coherence scores while exhibiting substantially weaker LLM-based
fluency/coherence and a large increase in detected mistakes. This apparent contradiction
is plausible given how these metrics operationalize ``coherence.'' Entity coherence in
TF3 is entropy-based and primarily reflects the \emph{balance and consistency of entity
distributions} within generated stories. Under aggressive quantization, generation often
becomes more stereotyped and repetitive (e.g., reusing a smaller set of entities and
reintroducing them with less variation). Such reduced diversity can increase distributional
stability and thus inflate entropy-based coherence, even as surface-level fluency and
Romanian grammaticality degrade. In other words, Q6 may become \emph{more uniform} in
entity usage while becoming \emph{less faithful} to natural Romanian syntax and phrasing.

\paragraph{Efficiency and deployment.}
Across compressed variants, the dominant gains are observed in storage footprint and
throughput. Quantization yields the largest storage reduction, whereas pruning-guided
distillation provides a complementary path that reduces model size while preserving stable
generation patterns through teacher supervision. Together, these results suggest that
moderate numerical compression (Q8) and distillation define two practical ``sweet spots''
for Romanian deployment, balancing efficiency against linguistic reliability.

\subsection{Comparison with Strong Instruction-Tuned Models}

To contextualize TF3 results relative to widely used instruction-tuned systems, we
evaluate several strong publicly available models under the same controlled prompting
and scoring setup. All models are evaluated on the same set of 100 Romanian synthetic
prompts using identical decoding parameters and the same LLM-as-a-judge rubric for
fluency and coherence.

Table~\ref{tab:instruction-baselines} reports average fluency, coherence, and mean
scores. Despite being trained entirely from scratch on synthetic data and operating
at a substantially smaller parameter scale, the TF3 Transformer and distilled student
remain within approximately 10\% of much larger, heavily instruction-tuned models.
This gap is modest given differences in model size, training data, and optimization
objectives, and highlights the strength of the TF3 pipeline under controlled
evaluation.

\begin{table}[!htbp]
\centering
\small
\begin{tabular}{lccc}
\toprule
Model & Fluency & Coherence & Mean Score \\
\midrule
LLaMA-3 8B Instruct & 4.70 & 4.72 & 4.71 \\
GPT-4.1-mini & 4.63 & 4.77 & 4.70 \\
DeepSeek (Instruct) & 4.71 & 4.66 & 4.68 \\
EuroLLM 9B Instruct & 4.58 & 4.66 & 4.62 \\
Gemma 2 9B Instruct & 4.46 & 4.58 & 4.52 \\
\midrule
TF3 Transformer (51.65M) & 4.28 & 4.09 & 4.19 \\
TF3 Distilled Student (26.45M) & 3.96 & 4.03 & 4.00 \\
\bottomrule
\end{tabular}
\caption{LLM-as-a-judge evaluation on 100 Romanian synthetic prompts. Scores are reported
on a 1--5 scale. TF3 models remain within approximately 10\% of strong instruction-tuned
baselines despite substantially smaller scale and purely synthetic training data.}
\label{tab:instruction-baselines}
\end{table}

\subsection{Qualitative Examples}
\label{sec:qualitative_examples}

To complement aggregate metrics, we inspect representative generations produced under
identical prompts by all evaluated models. Transformer-Q6 frequently exhibits reduced
lexical diversity and repetitive entity reuse, which can artificially increase
entropy-based entity coherence while degrading perceived fluency. Common failure modes
include agreement errors (gender and number), malformed diacritics or rare-word
distortions, and repetitive sentence templates that preserve high-level discourse
structure but sound unnatural at the sentence level. By contrast, the distilled
student generally maintains more stable syntactic patterns, with errors more often
reflecting simplified phrasing rather than structural breakdown.

\begin{quote}
\textbf{Example 1 (Transformer-Q6: agreement and fluency).} \emph{„Vulturul era mic și
puternic, dar nu îl lăsa să se ascundă. [...] Vulturul își dădu seama că nu era atât de
puternică pe cât crezuse.”}

\textbf{Example 2 (Transformer-Q6: repetition inflating entity coherence).} \emph{„Cioara
vicleană s-a gândit o clipă. [...] Cioara vicleană și-a împărtășit planul cu cioara
vicleană și i-a arătat o altă cale către comoara ei.”}

\textbf{Example 3 (Distilled Student: stable syntax, simplified style).} \emph{„O
căprioară curajoasă a hotărât să își ajute prietenii. A lucrat alături de celelalte
animale și a învățat că bunătatea aduce liniște și siguranță.”}
\end{quote}

\subsection{Summary}

Three trends emerge from Figure~\ref{fig:compression_results}.

\begin{enumerate}
    \item \textbf{The full Transformer teacher defines the strongest overall quality regime.}
    It provides the best intrinsic modeling performance among Transformer-family models and
    serves as the most stable reference point across grammatical and discourse-level metrics.

    \item \textbf{Moderate quantization provides large efficiency gains with limited behavioural change.}
    Transformer-Q8 preserves most of the teacher’s linguistic behaviour while substantially
    improving storage and throughput, making it a strong default choice when low-bit inference
    is acceptable.

    \item \textbf{Aggressive compression changes behaviour qualitatively, not only quantitatively.}
    Transformer-Q6 improves efficiency further but exhibits clear degradation in perceived fluency
    and grammatical reliability. At the same time, its unusually high entity-coherence score is
    consistent with more repetitive and distributionally uniform generations, which can inflate
    entropy-based coherence even as surface-level quality declines. The distilled student offers
    an alternative path: it reduces size substantially while preserving stable generation patterns
    through teacher supervision, yielding an attractive compromise for local deployment and
    Romanian-native dataset generation.
\end{enumerate}

Although TF3 does not include a formal large-scale human evaluation, representative
generations and evaluation outputs were manually inspected by Romanian-speaking
researchers with linguistic training to verify grammatical plausibility and
morphological correctness. These checks served as a qualitative validation layer
complementing automated metrics.

Overall, these results map a practical trade-off surface between capacity, quality, and
efficiency for Romanian language models trained on structurally controlled synthetic data,
informing model selection for both research and deployment scenarios.

\section{Discussion}
\label{sec:discussion}

\paragraph{Synthetic data as a viable foundation.}
The findings in Section~\ref{sec:results} demonstrate that carefully structured
synthetic corpora are sufficient to support the pretraining of compact Romanian
language models. Despite relying exclusively on machine-generated microfiction,
the 51.65M-parameter Transformer exhibits stable convergence, coherent Romanian
morphology, and strong predictive behavior. These results reaffirm a central
conclusion from TF1 \citep{nadas_tf1-en-3m_2025} and TF2
\citep{nadas_small_2025}: when narratives are produced through a controlled,
taxonomy-driven generative scaffold, synthetic text can serve as a reliable and
high-quality substrate for training language models in morphologically rich
languages.

\paragraph{Architectural efficiency and compactness.}
The experiments reveal that the 51.65M Transformer architecture provides a robust
balance between expressive capacity and computational efficiency. The structured
pruning analysis shows that substantial portions of its width and MLP capacity can
be removed with only moderate loss degradation, revealing meaningful redundancy.
By leveraging this pruning-informed insight, TF3 identifies an optimally compact
26.45M-parameter Transformer variant that maintains acceptable perplexity and
coherent generation while operating at a dramatically reduced memory footprint.
This emphasizes that intelligent, data-driven compression—rather than blind
downscaling—can produce compact Romanian LMs suitable for practical deployment.

\paragraph{Compression and the limits of reduction.}
Compression experiments further delineate the boundary between efficiency and
linguistic quality. Quantization to 8-bit precision preserves nearly the entirety
of the teacher's linguistic competence while reducing storage size by more than
half, making quantized models ideal for consumer deployment. More aggressive
methods—structured pruning and logit-based distillation—expose sharper
trade-offs. The distilled 26.45M student provides high throughput and compactness,
but exhibits reduced stylistic fluency and grammatical richness relative to the
full teacher. At the same time, some narrative metrics (such as entropy-based
entity coherence) show improvements, suggesting that distillation can regularize
discourse structure even as surface-level fluency weakens. Together with TF2's
observations on translation robustness, these results outline a practical Pareto
frontier for Romanian model deployment.

\paragraph{The value of multi-dimensional evaluation.}
The TF3 evaluation suite highlights the necessity of heterogeneous metrics for
inflectional languages. Intrinsic metrics such as perplexity and cross-entropy
capture predictive quality but fail to fully reflect narrative organization or
stylistic behavior. Rule-based grammar analysis identifies explicit morphological
violations, while the entropy-based entity coherence metric detects imbalances in
discourse structure. LLM-as-a-judge scoring provides a higher-level perspective
closer to human assessment. As in TF1 and TF2, these complementary modes confirm
that no single metric adequately characterizes the quality of compact Romanian LMs
and that multidimensional evaluation is essential for fair and reliable comparison.

\paragraph{Tokenizer design as a decisive factor.}
The experiments underscore the importance of language-specific tokenizer design for
morphologically rich languages. When evaluated under the full TF3 preprocessing and
training pipeline, the Romanian Unigram tokenizer does not provide substantial token-count
reductions relative to a Romanian-trained BPE tokenizer. Instead, its primary advantage
lies in producing more morphologically coherent and stable segmentations, particularly
with respect to stem preservation and productive inflectional suffixes.

While these properties do not necessarily minimize sequence length, they reduce brittle
or linguistically implausible subword splits and yield more consistent token boundaries
across inflected forms. This segmentation stability contributes to robust training
dynamics and supports morphosyntactic regularities during generation. In line with TF2’s
findings on the limitations of multilingual tokenization for EN--RO data, these results
position tokenizer design as a critical determinant of downstream modeling quality,
independent of raw token-count compression.

\paragraph{Answers to the research questions.}
Taken together, the empirical findings of TF3 provide clear and affirmative
answers to the research questions posed at the outset of this study.

\textbf{RQ1 (Tokenizer Design).}
Romanian-specific subword vocabularies address key limitations of generic multilingual
tokenizers by aligning segmentation with Romanian morphological structure. Under the
full TF3 preprocessing pipeline, a 32k Unigram tokenizer trained on Romanian microfiction
does not yield substantial token-count reductions relative to a Romanian-trained BPE
tokenizer, but provides more stable and morphologically faithful segmentations. In direct
comparison, Unigram better preserves productive stems and inflectional affixes, making it
well suited for Romanian language modeling under fixed context and capacity constraints.

\textbf{RQ2 (From-Scratch Pretraining).}
A compact 51.65M-parameter Transformer trained exclusively on synthetic moral
microfiction exhibits stable optimization dynamics, convergent perplexity, and
grammatically coherent Romanian generation. These results demonstrate that
carefully structured synthetic corpora are sufficient to support from-scratch
pretraining for morphologically rich languages, even in the absence of natural
text, provided that narrative structure and lexical diversity are controlled.

\textbf{RQ3 (Compression and Deployment).}
The base Romanian model admits substantial compression without catastrophic loss
of linguistic competence. While 8-bit quantization preserves nearly all predictive
and grammatical capacity at a fraction of the storage cost, more aggressive
compression techniques introduce sharper trade-offs. A pruning- and
distillation-informed 26.45M-parameter student model retains acceptable perplexity
and coherent discourse structure, though with some reduction in surface-level
fluency. These results delineate a practical Pareto frontier between efficiency
and linguistic quality, enabling informed deployment choices for resource-
constrained environments.

\paragraph{Ethical considerations and limitations.}
TF3 relies exclusively on synthetic data, avoiding privacy concerns associated with web-scraped corpora. However, synthetic-only pipelines may propagate biases present in the generator models, and the domain is restricted to moral microfiction. All artifacts are released under permissive licenses to support transparency and reproducibility. Future work should evaluate generalization to broader Romanian text domains and incorporate human evaluation to complement automated metrics.

\paragraph{}
Taken together, these observations establish TF3 as a demonstration of how
synthetic data, pruning-guided architecture design, Romanian-specific tokenization,
and multidimensional evaluation can be orchestrated to produce compact,
deployment-ready Romanian language models. The study extends the methodological
trajectory of TF1 \citep{nadas_tf1-en-3m_2025} and TF2 \citep{nadas_small_2025}
toward a fully mature, efficiency-centered development pipeline.

\section{Conclusion}
\label{sec:conclusion}

This paper introduced \textsc{TF3-RO-50M}, the third installment of the
TinyFabulist series and the first to train Romanian language models entirely from
scratch on a fully synthetic corpus. Building on the translation-based TF1 dataset
introduced in TF2 \citep{nadas_small_2025}, we constructed a large-scale Romanian
microfiction corpus, developed Romanian-specific subword tokenizers, implemented a
long-sequence preprocessing pipeline, and trained a 51.65M-parameter
LLaMA-style Transformer tailored to Romanian linguistic structure.

Our experiments support three central conclusions. First, synthetic corpora are a
viable and effective foundation for pretraining compact LLMs in morphologically
rich languages. The 51.65M teacher model achieves strong perplexity, stable
training dynamics, and coherent generation despite being trained exclusively on
synthetic text, validating and extending the TinyFabulist methodology introduced in
TF1 \citep{nadas_tf1-en-3m_2025} and refined in TF2 \citep{nadas_small_2025}.

Second, structured pruning reveals that the teacher contains substantial redundant
capacity. Guided by a pruning landscape that identifies the smallest viable
configuration under a 20--30\% loss increase, TF3 constructs a compact 26.45M
student model that preserves architectural depth while reducing width, attention
capacity, and MLP size. Knowledge distillation transfers the teacher’s probability
structure into this reduced model, resulting in a lightweight Transformer suitable
for CPU, mobile, and high-throughput inference environments.

Third, compression introduces meaningful but manageable trade-offs. Quantization
provides substantial memory savings at minimal quality loss, whereas deeper
distillation produces ultrafast but stylistically less expressive models. These
findings map a practical Pareto frontier for Romanian LMs, helping practitioners
select an appropriate balance between capacity, latency, and quality.

Beyond the modeling contributions, TF3 provides a comprehensive Romanian evaluation
framework integrating intrinsic metrics, morphological probes, entity coherence
analysis, rule-based grammar checking, and LLM-as-a-judge scoring. This suite
extends the reference-free evaluation paradigm of TF1 and TF2 and offers a robust
foundation for assessing compact models trained on synthetic corpora.

Future work will extend this pipeline by:
\begin{itemize}
    \item generating larger and more stylistically diverse synthetic corpora,
    \item exploring cross-lingual transfer between Romanian and other Romance languages,
    \item investigating hybrid architectures combining attention and structured state-space mechanisms,
    \item refining pruning and distillation techniques to better preserve stylistic richness.
\end{itemize}

Overall, TF3 demonstrates that high-quality Romanian language models can be trained
efficiently from scratch using synthetic data, and that compact Transformers remain
a promising direction for accessible, domain-adapted Romanian NLP systems.

\bibliographystyle{plainnat}
\bibliography{references}

\appendix

\section{Structured Pruning Framework}
\label{app:pruning}

This appendix details the structured pruning framework used in TF3 to estimate
the removable capacity of the pretrained 51.65M-parameter Romanian Transformer.
The framework is not proposed as a novel pruning method; rather, it serves as a
supporting empirical analysis that informs the design of the distilled student
model described in the main paper.

Retraining dozens of reduced architectures from scratch would be computationally
infeasible. Instead, TF3 adopts structured parameter masking, disabling controlled
fractions of MLP neurons and attention heads at inference time to simulate smaller
models. This approach draws on classical work in network pruning and
compressibility \citep{han_deep_2016, frankle_lottery_2019, michel_are_2019} and
enables rapid exploration of redundancy in a compact Romanian language model.

\subsection{Motivation and Principle}

Large Transformer models are known to contain redundant capacity, particularly in
their feedforward networks and multi-head attention blocks
\citep{michel_are_2019}. Early pruning research \citep{han_deep_2016} shows that
a large fraction of hidden units can be removed with only modest impact on
downstream performance, while subsequent work highlights the existence of highly
trainable sparse subnetworks \citep{frankle_lottery_2019}. 

TF3 extends these insights by applying structured parameter masking directly to the
pretrained 51.65M teacher model. By disabling controlled fractions of MLP neurons
and attention heads and measuring the resulting degradation in loss, the framework
provides an empirical estimate of the \emph{minimal viable capacity} for Romanian
narrative modeling. This avoids ad hoc architectural guessing and grounds the
student-model design in experimentally validated compression limits.

\subsection{Dimensions of Structural Downsizing}

The teacher model’s parameterization scales primarily along:

\begin{itemize}
    \item \textbf{Depth} (\texttt{num\_hidden\_layers}), determining the number of
    Transformer blocks.
    \item \textbf{Width} (\texttt{hidden\_size}), which governs the representational
    capacity of both self-attention and feedforward submodules.
    \item \textbf{MLP intermediate size} (\texttt{intermediate\_size}), the dominant
    contributor to total parameters in each block.
\end{itemize}

Among these, width and MLP intermediate size account for the largest share of
parameters. Prior work shows that pruning width often yields the greatest
compression gains \citep{han_deep_2016}. TF3 therefore focuses its pruning study
on the MLP and multi-head attention structures, which are empirically the most
redundant.

\subsection{Pruning Mechanisms}

TF3 applies two forms of structured pruning implemented via multiplicative masks
in the forward pass:

\paragraph{(1) Feedforward (MLP) Neuron Pruning.}
The intermediate projection of each block’s MLP holds a large proportion of the
model’s parameters. TF3 constructs binary masks over the intermediate activations,
zeroing out a specified fraction of neurons. Adjusting the pruning rate
effectively simulates narrower MLPs without modifying the underlying checkpoint or
retraining from scratch. From the perspective of the forward pass, pruned neurons
are permanently inactive and contribute neither activations nor gradients.

\paragraph{(2) Attention Head Pruning.}
Following structured head-pruning techniques \citep{michel_are_2019}, TF3 disables
entire attention heads by masking their output channels within the
query–key–value projections. Masking 30\%, 50\%, or 70\% of heads emulates
smaller multi-head configurations while preserving block depth and overall
architecture. Pruned heads no longer contribute to the attention mixture, which
directly reduces the model’s effective attention capacity.

\subsection{Methodological Procedure}

The pruning evaluation proceeds as follows:

\begin{enumerate}
    \item Load the pretrained 51.65M-parameter teacher checkpoint.
    \item Construct MLP and attention-head masks corresponding to chosen pruning
    rates.
    \item Apply these masks during forward passes of the model to disable the
    selected neurons and heads.
    \item Evaluate cross-entropy and perplexity on a held-out set of Romanian
    narratives.
    \item Sweep across a two-dimensional pruning grid to generate a loss and
    perplexity landscape as a function of MLP and attention pruning rates.
\end{enumerate}

This mirrors compressibility studies in prior pruning literature
\citep{han_deep_2016, michel_are_2019}, while adapting the methodology to a
compact Romanian LLaMA-style model.

\subsection{Selecting the Target Student Capacity}

The pruning grid reveals that substantial fractions of the teacher model may be
disabled with only moderate performance degradation. TF3 selects the maximal
pruning configuration that remains under a 20--30\% loss increase—specifically,
\textbf{50\% MLP pruning and 30\% attention-head pruning}, corresponding to a
measured degradation of roughly 26--27\%. This configuration defines the target
architecture for the distilled student model: reduced-width MLPs, fewer attention
heads, and tied embeddings.

This data-driven procedure ensures that the student inherits the smallest
structural capacity capable of maintaining acceptable Romanian modeling quality,
providing a principled foundation for the knowledge-distillation process described
in Section~\ref{sec:distillation}.

\section{LLM-as-a-Judge Evaluation Prompt}
\label{app:llm-judge-prompt}

The following prompt template was used for Gemini 2.5 Flash–based fluency and coherence evaluation, accessed via OpenRouter. The model receives the generated Romanian text and returns structured JSON output for automated aggregation. We include both the system instruction and the user prompt as used in our implementation.

\begin{verbatim}
system = "You are an expert in Romanian language. You evaluate texts for fluency, coherence, and
grammatical accuracy with precision. Always respond in valid JSON format only, no additional text."

prompt = f"""Evaluate the following Romanian text line for fluency, coherence, and grammatical mistakes.

Text:
{text}

Analyze the text and:
1. **Fluency** (0-100): How natural and fluent the text sounds in Romanian. Consider:
   - Natural word order and phrasing
   - Appropriate use of Romanian expressions
   - Smooth flow and readability
   - 100 = perfectly fluent, natural Romanian
   - 80-99 = very fluent with minor awkwardness
   - 60-79 = mostly fluent but some awkward phrasing
   - 40-59 = somewhat fluent but noticeable issues
   - 0-39 = not fluent, very awkward

2. **Coherence** (0-100): How well the text makes sense and maintains logical flow. Consider:
   - Logical structure and meaning
   - Clear connections between ideas
   - Consistency in narrative/argument
   - 100 = perfectly coherent and clear
   - 80-99 = very coherent with minor issues
   - 60-79 = mostly coherent but some confusion
   - 40-59 = somewhat coherent but unclear parts
   - 0-39 = incoherent, confusing

3. **Mistakes**: Count and identify all grammatical mistakes (agreement, conjugation, declension, word choice, etc.)

Respond in JSON format with the following structure:
{{
  "fluency": <score 0-100>,
  "coherence": <score 0-100>,
  "total_mistakes": <exact count of mistakes>,
  "mistakes": [
    {{
      "position": "<position in text>",
      "type": "<type of mistake>",
      "original": "<incorrect text>",
      "correction": "<suggested correction>",
      "explanation": "<short explanation>"
    }}
  ]
}}"""
\end{verbatim}

Each dimension (fluency, coherence, grammatical correctness) is scored on a 0--100 scale. Grammatical errors are extracted with suggested corrections to enable fine-grained error analysis.

\end{document}